\documentclass[preprint,12pt]{elsarticle}

\usepackage[utf8]{inputenc}
\usepackage{amssymb}
\usepackage{rotating}
\usepackage{multirow}
\usepackage{amsmath}
\usepackage[table]{xcolor}
\usepackage{graphicx}
\newtheorem{definition}{Definition}

\journal{Knowledge-Based Systems}

\begin{document}
\begin{frontmatter}
%opening
\title{An artificial intelligence tool for heterogeneous team formation in the classroom}
\author[upv]{Juan M. Alberola}
\author[upv]{Elena del Val}
\author[cov]{Victor Sanchez-Anguix}
\author[upv]{Alberto Palomares}
\author[upv]{Maria Dolores Teruel}
\address[upv]{Universitat Politècnica de València, Cami de Vera s/n, 46022, Valencia, Spain}
\address[cov]{Coventry University, Department of Computing, Gulson Road, CV1 2JH}

\begin{abstract}
Nowadays, there is increasing interest in the development of teamwork skills in the educational context. This growing interest is motivated by its pedagogical effectiveness and the fact that, in labour contexts, enterprises organize their employees in teams to carry out complex projects. Despite its crucial importance in the classroom and industry, there is a lack of support for the team formation process. Not only do many factors influence team performance, but the problem becomes exponentially costly if teams are to be optimized. In this article, we propose a tool whose aim it is to cover such a gap. It combines artificial intelligence techniques such as coalition structure generation, Bayesian learning, and Belbin's role theory to facilitate the generation of working groups in an educational context. This tool improves current state of the art proposals in three ways: i) it takes into account the feedback of other teammates in order to establish the most predominant role of a student instead of self-perception questionnaires; ii) it handles uncertainty with regard to each student's predominant team role; iii) it is iterative since it considers information from several interactions in order to improve the estimation of role assignments. We tested the performance of the proposed tool in an experiment involving students that took part in three different team activities. The experiments suggest that the proposed tool is able to improve different teamwork aspects such as team dynamics and student satisfaction. 
\end{abstract}

\end{frontmatter}

\section{Introduction}
\label{sec:intro}
In the last few years there has been increasing interest in teamwork skills in the area of Higher Education \cite{dunne00,adams03,alexander06,fredrick08,sancho09}. Many plans of study and faculties have included general teamwork competence as a part of their educational programs for undergraduate students. The reasons for this inclusion are well-grounded in its pedagogical effectiveness and our current industrial paradigm. Firstly, the area of collaborative learning, supported by computers, promotes collaboration and makes learning more effective \cite{stahl2006group}. Secondly, the industry has shifted from an individually oriented work environment towards a team-oriented workplace. Nowadays, teams are at the heart of a vast majority of modern companies \cite{tarricone02,van03,fay06,ratcheva09}. Despite the often difficult decision-making tasks involving groups of individuals \cite{kerr04,behfar08}, teams have proven to have an inherent ability to solve the complex problems that are confronted in the current work environment. 

Given this context, it is fairly reasonable for Higher Education institutions to place a special emphasis on teamwork skills as a part of every program's learning outcomes. Unfortunately, not every single team is successful in their goals, and many teams fail due to incorrect team dynamics, lack of communication, and interpersonal conflict among team members \cite{de03,behfar08,parker11}. Even though some of the aforementioned problems can be alleviated with teamwork experience, these negative factors should be avoided whenever possible as they may generate resentment towards teamwork. Hence, identifying the patterns that drive successful teams and forming work teams according to these patterns become crucial tasks for every organization. Classrooms are not immune to this issue (specially if students are to learn teamwork skills), and unnecessary problems may hinder this learning process.

One of the most important theories regarding successful team dynamics is Belbin's role taxonomy \cite{belbin96}. In this theory, Belbin identifies eight heterogeneous behavioural patterns that are present in many successful teams in the industry: \emph{plant, resource investigator, coordinator, shaper, monitor evaluator, team worker, implementer, and finisher}. These behavioural patterns (or roles) should be played by the different team members in order to facilitate successful teamwork. Belbin's taxonomy has given rise to a wide variety of studies showing the theory's strengths and weaknesses \cite{fisher96,senior97,partington99,aritzeta07,belbin11}, it has been applied to a wide variety of domains \cite{Sommerville,smith00,rajendran05,king06}.

%Due to these reasons, the classroom environment would benefit from the application of Belbin's theory. Yet, there are several problematic circumstances that should be addressed for applying Belbin's role taxonomy to the classroom: i) Belbin's roles are classically identified by means of self-perception questionnaires. However, self-perception with respect one's own behaviour in a team may differ from those patterns perceived by teammates in a real team environment \cite{senior98}. ii) Individuals are not purely described by strict roles, and one person may show patterns which make uncertain the most prominent role of the individual. iii) Even a small classroom with 30 students has ${30 \choose 5} = 142506$ different teams of five individuals, and the total of team configurations for the classroom explodes exponentially with this quantity. Finding the best possible configuration poses a computationally expensive problem for it to be solved manually.

As shown by several studies, the classroom environment may benefit from the application of Belbin's theory \cite{blignaut98,henry99,van08,ounnas09,yannibelli2012a}. One of the reasons for this successful application in education is the identification of behavioral patterns that are present in many group dynamics. However, there are several problematic circumstances that should be addressed in order to apply Belbin's role taxonomy to the classroom. The first one is that Belbin's roles are classically identified by means of questionnaires (mainly self-perception questionnaires) that are filled out before working with others. However, self-perception results %with respect one's own behaviour in a team 
may differ from those patterns shown in a real team environment \cite{senior98}. Therefore, we believe that a more effective role assignment could be achieved by considering both the information collected before working on a team and the feedback provided from peers after working on a team. 
%De l'altre paper: Second, our pro posal provides a more reliable role assignment since it considers the opinion of other members instead of a personal evaluation. Finally, in each iteration of the algorithm, the solution is improved with the feedback received from direct interactions among students in each team.
The second one is that individuals are not purely described by just a static and strict role. Despite the fact that, due to the individual's personality, one may have a most predominant role, individuals show a rich variety of behavioral patterns depending on circumstances. Firstly, this makes the most prominent role of the individual uncertain, as the individual may show a range behaviors for different scenarios. Secondly, as the individual may show different behavioral patterns, the individual behavior may be best described as a probability distribution over such patterns or roles that he/she plays. The third one is that even a small classroom with 30 students has ${30 \choose 5} = 142506$ different teams of five individuals, and the total number of team configurations for the classroom explodes exponentially with this amount. Finding the best possible configuration poses a computationally expensive problem for it to be solved manually.

In this article, we present a computational tool that attempts to address the aforementioned problems. The tool is based on artificial intelligence (AI) and iterative interactions/feedback. The use of AI techniques allows us to address uncertainty and solve computationally expensive problems. More specifically, the tool makes use of Bayesian learning to tackle uncertainty with regard to students' prominent roles, and the problem of finding optimal teams is treated as a coalitional structure generation problem \cite{rahwan15}, which is solved by means of linear programming methods. Additionally, the proposed tool is iterative in nature: it proposes team configurations for class task assignments and then it gathers feedback from team members with respect to the roles portrayed by the other teammates. This information is later used to refine future team configurations proposed by the tool.

The remainder of this article is organized as follows. Section \ref{sec:belbin} describes the main features of the Belbin model. Section \ref{sec:wkflow} presents how the tool would generally work and some implementation details. Section \ref{sec:team} preswents an in-depth explanation of the mechanism used for team formation, which is at the core of our team formation tool. Section \ref{sec:exp} analyzes the impact of testing our proposal in a real educational environment. Section \ref{sec:related} shows the most relevant works in the literature with regard to team formation tools. Finally, Section \ref{sec:con} presents some concluding remarks and future work.

%\label{sec:belbin}
\section{The Belbin theory}\label{sec:belbin}
Prior to detailing how the proposed tool was implemented, we believe that it is important for the reader to be familiar with the Belbin theory since it is one of the fundamental pillars of our tool. The Belbin theory \cite{belbin96,Belbin1993,belbin2009belbin,Belbin2011} provides a thorough of the influence of different types of roles in teamwork. A team role is defined as a behavioral pattern that facilitates the progress of the whole team. Assuming that there would probably be boundless behaviour patterns, Belbin states that the range of behaviours that really influence the performance of a team is limited. In Belbin's model, a role is defined by six factors: personality, mental ability, current values and motivation, field constraints, experience, and role learning \cite{aritzeta07}. Specifically, Belbin defines the following eight roles:

\begin{itemize}
\item \textbf{Plant/Creative}: is creative and imaginative. He/she generates ideas and solves difficult problems.
\item \textbf{Resource investigator}: is outgoing and communicative. He/she explores opportunities and interacts with people outside the team.
\item \textbf{Co-ordinator}: is mature and confident. He/she has a global view of the project and delegates effectively. 
\item \textbf{Shaper}: is challenging and dynamic. He/she has the drive and courage to overcome obstacles.
\item \textbf{Monitor evaluator}: is sober, strategic, and discerning. He/she sees all options and judges accurately.
\item \textbf{Teamworker}: is co-operative, perceptive, and diplomatic. He/she is able to listen and avert friction.
\item \textbf{Implementer}: is practical, reliable, and efficient. He/she turns ideas into actions and organizes work that needs to be done.
\item \textbf{Completer finisher}: is painstaking, conscientious, and anxious. He/she searches out errors, polishes, and perfects them.
\end{itemize}

In later revisions of this theory, a ninth role of \emph{specialist} was introduced for the case when technical expertise is necessary for the performance of certain tasks. Belbin's model has been associated to behaviours and performance. In line with other authors, Belbin has argued that the most successful teams are composed of a balanced combination of the above roles, ideally all of them. In contrast, teams composed of homogeneous roles tend to provide unsatisfactory results. 

The Belbin model is traditionally operationalized through the Team Role Self-Perception Inventory, which allows each individual to discover his/her most prominent role based on his/her own judgment. The main disadvantage of this self-perception questionnaire is that individuals may have a preconceived image of themselves, which is diametrically different to the image that is reflected to others \cite{john94,paulhus98,senior98,pronin04}. Complementary to this, an Observer Assessment Sheet was also designed to be used by other colleagues who could make an informed judgment based on their knowledge of an individual. However, this questionnaire usually assumed that the observer should know the individual that was being evaluated in depth. This is something that is not always possible to assume in higher education contexts. 

%In the following sections we present our proposal that is aimed at obtaining the most promising teamwork composition based on the Belbin model. This computational-based solution obtains the teams' composition in a reasonable response time that could not be solved manually. In contrast to other approaches, we do not establish the role of each individual before running the process by means of questionnaires. The most prominent role of each individual is determined by the rest of his/her partners after working with them.  Thus, as the information used to characterize each individual is collected from others and this increase as the number of iterations increase, the role associated to each individual can be more accurately determined.

%\label{sec:wkflow}
\section{General tool workflow}\label{sec:wkflow}
In this section, we describe the general workflow of our tool and its most important features.  During an academic course, a teacher may carry out several team activities that require the formation of teams. As mentioned above, one of the main problems for teachers is to optimally create teams when there is no previous information about student profiles, and the number of students is high. In the latter scenario, the complexity of determining optimal teams is complex for the teacher due to the exponential nature of the problem \cite{hansen06}. In order to provide support for this team management task, we have developed a software application for teachers that facilitates the costly task of dividing students into optimal or near optimal teams. As a general outline, the application relies on student feedback, coalitional structure generation, and Bayesian learning to form proper distributions of student teams. In the following paragraphs, we will explain how these elements are put together to provide an adequate team 
formation tool.

\begin{figure}[t]
\begin{center}

\includegraphics[width=1\linewidth]{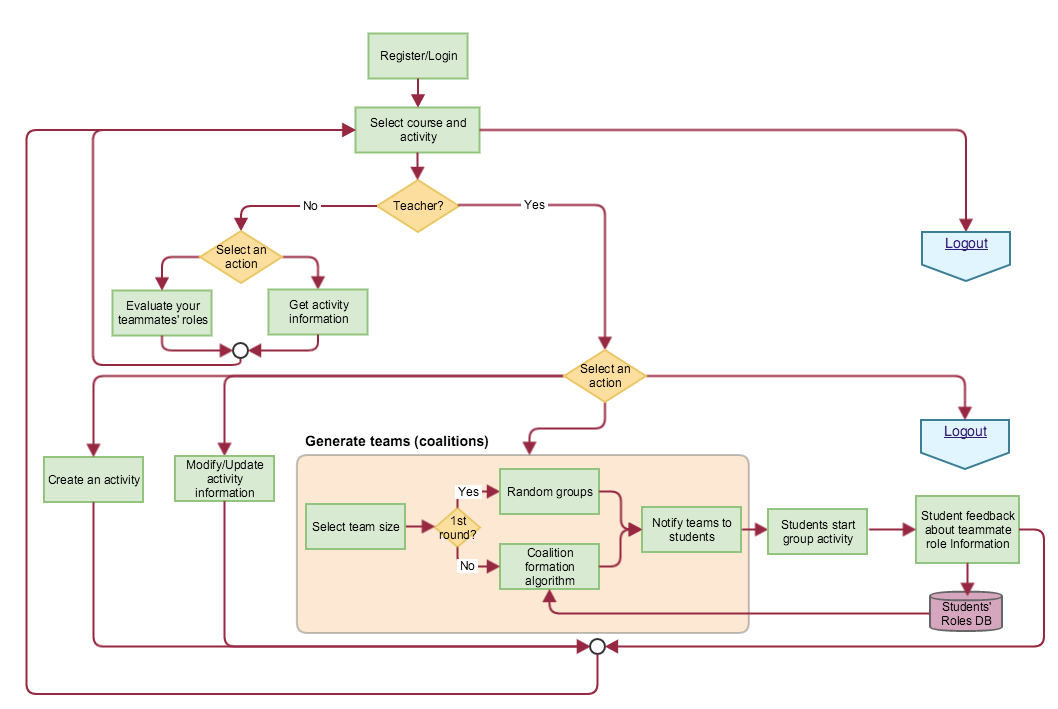}
\caption{Activity diagram flow for the team formation tool}
\label{img:activityDiagram}
\end{center}
\end{figure}

\begin{figure}[t]
\begin{center}

\includegraphics[width=0.5\linewidth]{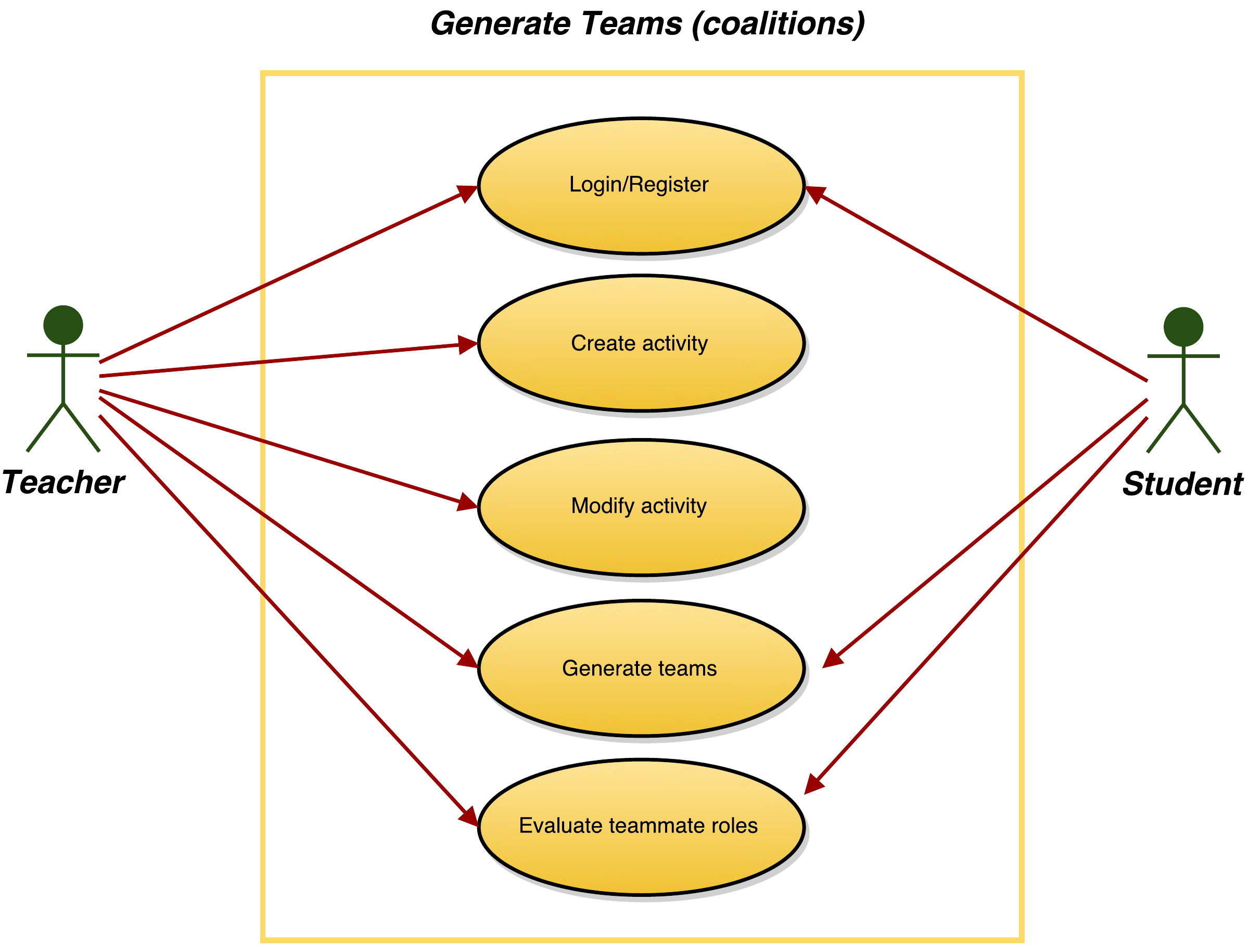}
\caption{Use cases}
\label{img:useCase}
\end{center}
\end{figure}

It should be noted that the tool has been designed to be integrated in web platforms where the actors (i.e., teachers and students) can interact with the system. We have a standalone web application, and we have also worked towards the integration of part of the tool's functionality into Sakai\footnote{http://sakaiproject.org}, which is the e-learning platform of choice at the Universitat Politècnica de València. The main functionalities of the system can be observed in the UML use case diagram shown in Figure \ref{img:useCase}. A more detailed diagram flow can be observed in Figure \ref{img:activityDiagram}.

\begin{figure}[t]
\centering
\includegraphics[scale=0.4]{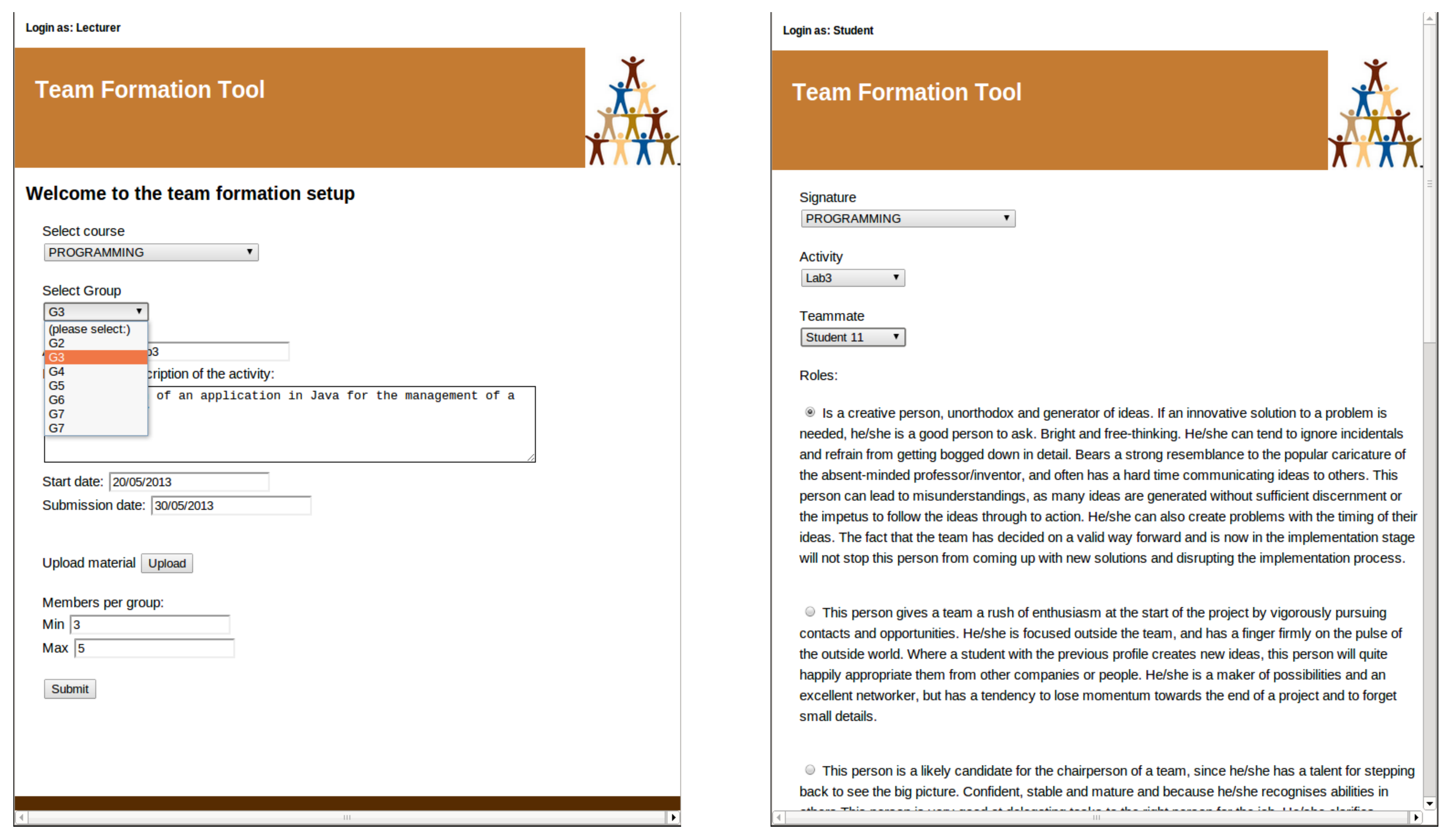}
\caption{(Left) Activity creation form. (Right) Teammate evaluation web form in our standalone application.}
\label{webPAAMS}
\end{figure}

The initial starting point for the application usually corresponds to the teacher login in the system (see Figure \ref{webPAAMS}, left). The teacher can then create a new team activity. After that, the web application shows the associated modules to the teacher in a pull-down menu. The teacher chooses the module where the team activity will be developed. The teacher can also fill out all the fields associated to the activity (activity description, start date, end date, on-line material for the activity). The teacher should then determine the maximum and minimum size for the student teams. 

Taking into account these parameters, the team formation mechanism is ready to generate an automatic proposal of teams. If it is the first time that the tool takes a particular group of students as input, the tool does not have information about the students' roles in the database. Therefore, the first set of teams is generated randomly. Otherwise, the system may employ a more complex team formation algorithm, taking into account the information provided by students in previous team activities. In this article, we employ a team formation algorithm that is based on Belbin's theory, student feedback, coalitional structure generation, and Bayesian learning. This algorithm is described in detail in Section \ref{sec:teamformation}. However, we have designed a modular, loosely-coupled web tool where each functionality is provided by a different software module in order to facilitate adaptability to new requirements or modifications. Hence, it is easy to adapt the tool to consider other criteria for group generation (e.g., personality traits, students marks, etc.). Once the teams have been formed, they are sent via email to the teacher, who can in turn notify students about their teammates. The teacher can then publish the team configuration.

\begin{figure}[t]
\centering
\includegraphics[scale=0.3]{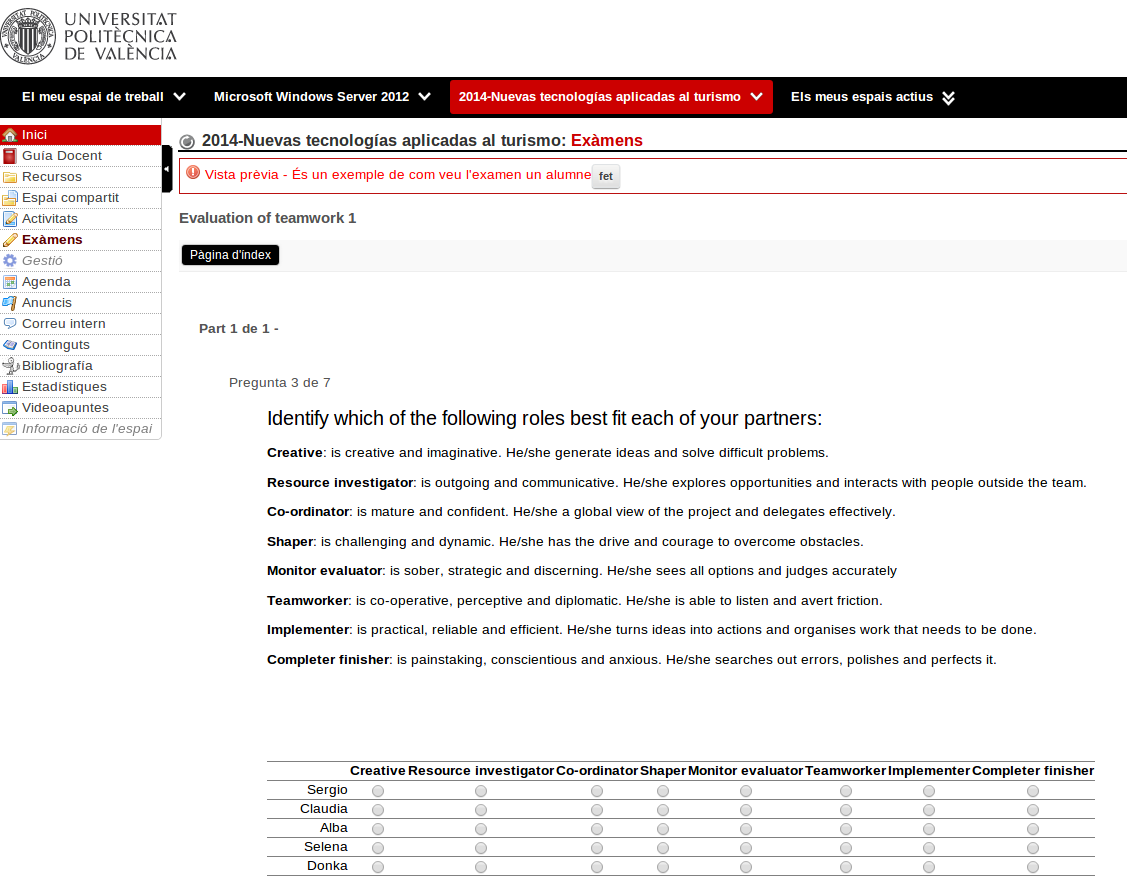}
\caption{Team evaluation web form in Sakai}
\label{Poliformat}
\end{figure}

Once the activity has been carried out in the classroom, each student receives a notification from the system to provide feedback about his/her teammates according to Belbin's roles. Student feedback is one of the main aspect of our team formation tool because this information will be used in future team formation tasks. In order to provide feedback, each student should log in the application. Then, the student chooses the module, the activity, and the teammate to be evaluated (see Figure \ref{webPAAMS}, right). In contrast to other Belbin approaches, we have advocated for a simple feedback questionnaire. The reason for this is that students do not get to know their team members well enough during team activities to be able to fill out the Observer Assessment sheet \cite{senior1998comparison} (i.e., a questionnaire used by colleagues of the individual or those who know him/her well). The application shows the description of the roles, and, at that point, the student should assign a role description to each teammate. As the reader can observe, we are using students as classifiers, and the performance of the tool is linked to the ability of humans for such a task. The information is then gathered by the application and stored in a database. Besides the information about the roles of his/her teammates, the student fills out a form with information about the degree of satisfaction with the team and with the methodology. The evaluation of the role of each team member has also been integrated in Sakai (see Figure \ref{Poliformat}). This integration facilitates the provision of the data by the students since the feedback form is available in the module of the course. As new team activities are carried out, new information is gathered and accumulated so that this information can be used in future team formation tasks. The idea behind this mechanism is that as the system gets more information, the system would have more evidence about the predominant roles of each student (i.e., the most often observed behavioral patterns) and therefore better team allocations can be provided. 

%\label{sec:team}
\section{Team formation mechanism}\label{sec:team}
\label{sec:teamformation}
At this point, we have described the general use of the tool. In this section, we describe our policy for dividing students into teams. This policy relies on student feedback, coalitional structure generation, and Bayesian learning to form proper distributions of student teams. First, we describe how dividing students into optimal teams is equivalent to a coalition structure generation problem. Then, we describe how Bayesian learning is employed to update the information of the classroom and to determine the most relevant role of each student.

\subsection{Student team formation as a coalition structure generation problem}
The Coalition Structure Generation problem refers to partitioning the components of a set into exhaustive and disjoint coalitions so that the global benefits of the system are optimized. In our problem, the components of the set are the students that take part in a classroom team activity:

\begin{definition}
	Let $A = \{a_i, ..., a_n\}$ be a set of students, and let $R=\{r_1,r_2,...,r_m\}$ be the set of roles that a student may play (in our case, Belbin's roles), and let $role_{i}$ denote the most predominant role of $a_{i}$. Let $T_j\in A$ be a subset of $A$ called \textit{team}.
	\end{definition}

The value of a team $T_j$ is given by a characteristic function $v(T_{j})$. A characteristic function $v(T_{j}) : 2^A \rightarrow \mathbb{R}$ assigns a real-valued payoff to each team $T_j$. The value of a team $v(T_{j})$ is calculated on the basis of the most predominant role that each student $a_{i} \in T_{j}$ has ($role_i$). Let $k$ denote the size of the team $T_{j}$ and $\pi_{j}=\{role_{1},...,role_{k}\}$ with $\forall r'_{i} \in R$  be a vector with the most predominant role of each team member of $T_{j}$. In that case, $v(T_{j}) = v(\pi_{j})$.  

Unfortunately, it is not possible to know the main role of each team member $\pi_{j}$ exactly and therefore $v(\pi_{j})$ cannot be calculated precisely. However, it is possible for us to calculate an estimation of the value of the team $v(T_j)$ given the history of evaluations received for each student $H$ that is gathered and accumulated after each team activity. Let  $\pi_{j}'=\{role_{1}=r'_{1},...,role_{k}=r'_{k}\}$ be a vector containing a set of hypotheses for the most predominant role of each team member, and let $\Pi$ be the set of all possible vectors of hypotheses for predominant roles of $T_{j}$. We calculate the expected value of a team given the history of evaluations as: 

\begin{equation}
\hat{v}(T_{j} | H) = \underset{\pi' \in \Pi}{\sum} p(\pi'|H)\times v(\pi') =  \underset{\pi' \in \Pi}{\sum} \left(v(\pi') \times \underset{a_{i} \in T_{j}}{\prod} p(role_{i}=r'_{i} | H)\right)
\label{expectedteam}
\end{equation}
where $p(\pi'|H)$ represents the probability for $\pi'$ to be the real role distribution in $T_{j}$ given the history of evaluations $H$. Each $p(\pi'|H)$ can be split into its $p(role_{i}=r'_{i} | H)$ since we assume that the role of each student is conditionally independent of other students' roles given the history of evaluations.

Conforming to different studies \cite{Higgs05}, the team should take advantage of having a balanced distribution of roles (i.e., one person per role) and is expected to obtain better results than a homogeneous team. In the coalition generation problem presented in this article, the overall effectiveness of each team $v(\pi')$ is measured numerically depending on the distribution of roles in the team as follows: 

\begin{equation}
v(\pi')=\frac{1}{2^{D(\pi')}}, 
\end{equation}\label{eficacia}
\noindent where $D(\pi')$ calculates the number of roles that are repeated in $\pi'$.

This function exponentially penalizes repeated roles in a team. For instance, a team that consists of four students all of whom have distinct roles will have a maximum efficiency of 1 (the best team configuration); a team of four where there are two students with the same role will have an efficiency of 0.5 (this value tries to reflect that temporary arguments might arise in the team where students with equal roles appear); a team where there are three students with the same role will have an efficiency of 0.25 (this value reflects that there is a higher probability that many problems will arise on the team due to the high number of students with a similar profile and the efficiency of the team would be affected); and finally, a team where the four students have the same dominant role will have an efficiency of 0.125.

Once we have defined how a team is created and its associated value, we define the concept of \textit{team structure}. 
\begin{definition}
A \textit{team structure} $S=\{T_1,T_2,...,T_k\}$ is a partition of teams such that $\forall i,j (i \neq j), T_j\cap T_i = \emptyset, \underset{\forall T_j \in S}{\bigcup} T_j = A$. 
\end{definition}
The value of a team structure is denoted by $v(S)$, where $v(S)$ is an evaluation function for the team structure. 
In this work, we consider that the quality of each team is independent of other teams. Therefore, we can calculate the value of the team structure as $v(S) = \underset{T_{j} \in S}{\sum} v(T_{j})$. 

The goal of the application is to determine an optimal team structure for the classroom $\underset{S \in 2^{A}}{argmax} \;\;v(S)$. Taking into account Equation \ref{expectedteam} to calculate the expected value of a given team, our team formation problem requires that the following expression be solved at each iteration:
\begin{equation}
\underset{S \in 2^{A}}{argmax}\;\; \underset{T \in S}{\sum} \hat{v}(T | H) 
\label{expectedstructure}
\end{equation}

It turns out that partitioning a set of students into disjoint teams while optimizing a social welfare function corresponds to the formalization of coalition structure generation problems. In order to solve this problem, we formally define the coalition structure generation problem as a linear programming problem \cite{ohta09} and solve it with the commercial software \textit{ILOG CPLEX 12.5} \footnote{http://www.ibm.com/software/commerce/optimization/cplex-optimizer/}.

\subsection{Estimating students' roles via Bayesian learning}

After every activity, each student evaluates his/her peers by stating the most predominant role of each of his/her teammates. Therefore, new information about the most predominant role of each student is available and the history of evaluations, $H$, grows. Hence, at each iteration, the application updates information regarding the probability for a student $a_{i}$ to have $r'_{i}$ as his/her most relevant role given the evaluation history $p(role_{i}=r'_{i}| H)$. We employ Bayesian learning for this matter:

\begin{equation}
p(role_{i} = r'_{i} | H ) = \frac{p(H|role_{i}=r'_{i})\times p(role_{i}=r'_{i})}{\underset{r \in R}{\sum} p(H|role_{i} = r) \times p(role_{i}=r)}
\end{equation}
where $p(H|role_{i}=r'_{i})$ is the likelihood function and $p(role_{i}=r'_{i})$ is the prior probability for the hypothesis. For the likelihood function, we can calculate it as $p(H|role_{i}=r'_{i}) = \frac{\#\{r'_{i} \in H_{i}\}}{|H_{i}|}$, where $H_{i}$ denotes the peer evaluations about agent $a_{i}$ and $\#\{r'_{i} \in H_{i}\}$ indicates the number of times that $r'_{i}$ appears as evaluation in $H_{i}$. As for the prior probability, we calculate it as $p(role_{i}=r'_{i})=\frac{\#\{r'_{i} \in H\}}{|H|}$. Laplace smoothing \cite{russell10} is employed to ensure that the likelihood for each role hypothesis can be calculated in the first iterations. 

This way, we build a probability distribution for each student over the roles that he/she plays. This mechanism allows us to model the fact that individuals are complex, and they may show a variety of behavioral patterns (although one of them may be the most predominant one). In Equation \ref{expectedteam} we use this information to estimate the expected value for a team given the probability distribution of students over behavioral patterns.  

%\label{sec:exp}
\section{Experiments}\label{sec:exp}
This tool was validated in an experimental setting involving real students. More specifically, the experiment was carried out in two consecutive modules in the \textit{Tourism Degree Program of} the \textit{Universitat Politècnica de València} during the academic year 2013/2014. The first module is held during the first semester, whereas the second is held during the second semester. Both modules are compulsory for students, and they are placed in the same academic course. Therefore, students attending the first module usually also attend the second module. 

There may be some students that only participate in one of the two modules (i.e., the first or the second module). However, one of our goals is to check the robustness of the proposed tool in a realistic setting where the tool has less information regarding some of the students. The reason for selecting these two modules is that they have been classically taught using cooperative learning methods. 

\subsection{Experimental setting}
The experiment was divided into three team activities that were split between the two modules. The first activity performed in the first module, while the second and third team activities performed in the second module. Prior to the start of each team activity, the tool proposes a set of teams ranging from 4 to 6 members based on the available information. For the preparation of the first team activity, since there is no information about students, the students are randomly placed into teams. We advocated for the use of random teams instead of letting the students decide their own teams. %due to the fact that it is a more realistic setting in the industry. Very frequently, professionals do not decide who they work with  and they have to adapt and learn to work with unknown colleagues. We believe that this is an important lesson for students and, thus, we decided not to allow students to initially decide their teams.

A total of 77 students participated at some point in the experiment. Of that set of students, 60 students participated in the first team activity and 50 students participated in the second and third team activities. A total of 34 students participated in all of the team activities, while 19 additional students participated in the second and third team activities. It should be noted that the first team activity acts as the benchmark for team performance, and the second and third team activities reflect the effect of the proposed tool on team performance. Even in the case of the second activity, where 19 additional students took part in the activity, a majority of the students (i.e., 34 students) already had peer evaluation from the first activity.

Apart from the peer evaluation that is carried out after each team activity, we devised a questionnaire that contains questions about different teamwork aspects such as communication, coordination, team dynamics, etc. These questionnaires were answered by students at the end of each team activity, and, in general, they could reflect different levels of satisfaction using Likert scales. The questions have been organized according to two main categories: team dynamics (i.e., coordination, cooperation, work balance, team member behaviour, decision making, etc.) and student satisfaction. The specific questions can be found in Table \ref{tab:question}. 

\begin{sidewaystable}
\caption{The questionnaire handed to students after each team activity}
\begin{center}
\begin{tabular}{| l | p{4cm} |} 
\hline
\multicolumn{2}{|c|}{\textbf{Team dynamics}}\\
\hline
Coordination \& Cooperation & \\
Q1. Our team has been well-coordinated & Likert scale \\
Q2. Our team created a cooperative environment & Likert scale \\
Q3. We established specific norms and tasks for everyone & Likert scale \\
Q4. Each member did his/her share of work & Likert scale \\
Q5. Our team spent its time well & Likert scale \\
Team member behaviour & \\
Q6. The attitudes of team members were positive & Likert scale \\
Q7. Team members treat each other in a respectful way & Likert scale \\
Decision making & \\
Q8. Team members accepted suggestions in a positive way & Likert scale \\
Q9. Our team listened to the opinions of all the members & Likert scale \\
Q10. It was difficult for my team to come to a decision & Likert scale \\
\hline
\multicolumn{2}{|c|}{\textbf{Student satistaction}}\\
\hline
Q11. How was your experience on this team? & Likert scale\\
Q12. I felt comfortable working on my team & Likert scale \\
Q13. I felt that I contributed to this team & Likert scale \\
Q14. Our project is acceptable to everyone on the team & Likert scale \\
Q15. Evaluate your team with a grade from 0 to 10 & 0 to 10 grade \\
\hline
\end{tabular}
\end{center}
\label{tab:question}
\end{sidewaystable}%

\subsection{Results}
In order to analyze the performance of the proposed tool regarding satisfaction at the team and the individual levels, we compared the questionnaires from the first team activity versus the questionnaires obtained after the second and theard team activities. To analyze data from Likert scales, we first combined categories into binary categories (i.e., disagree or indifferent/agree) since some of the options did not have enough samples to generalize (i.e., less than 5 counts). 

When combined into binary categories, the result for each question are two 2x2 contingency tables, one for the comparison between the first team activity and the second team activity and one for the comparison between the first team activity and the third team activity. There are two significant problems when studying team formation in the classroom. On the one hand, many classes are not composed of a large number of students, and, therefore, the number of samples per experiment tends to be low and it is difficult to include more samples. In this sense, the setting resembles that of the life and medical sciences. On the other hand, it has been reported in the literature that, when the studied variable is discrete (as in our contingency tables), classic calculations of the p-value do not represent its classic meaning \cite{hirji91,agresti98,newcombe98,hwang01,agresti11,graffelman13}. Due to the discrete nature of the variable, only one set of p-values is possible and the method tends to be excessively 
conservative and far from the meaning of classic p-values. 
For these scenarios, researchers propose the calculation of the mid p-value, whose type I error rate is closer to the nominal level. We employed a one-tailed test ($\alpha=0.05$) to test if the ratio of positive answers in the second activity is greater than the ratio of positive answers in the initial activity \cite{jewell03,rothman08}. For these tests, we calculated the mid p-value since its type I error rate is closer to the nominal level. Similarly, we carried out another one-sided proportion test ($\alpha=0.05$) to test if the proportion of positive answers in the final activity is greater than the proportion of positive answers in the initial activity. In order to account for possible type I errors, we adjusted the mid p-values obtained using Hommel's method. 

As for numeric answers (Q15), a one sided Mann Whitney statistical test ($\alpha=0.05$) was employed to assess statistical differences between the null hypothesis and alternatives, with the null hypothesis representing the same underlying distribution, and the alternative hypothesis representing an underlying distribution whose values are larger than those presented by the other distribution. In order to account for possible type I errors, we adjusted obtained p-values by using Hommel's method. 

The results for the experiment can be viewed in Table \ref{tab:result}. The table shows (in bold font and green shadow cell) those scenarios where the alternative hypothesis was accepted with 95\% confidence and it held significantly better results than those gathered in the initial activity. A blue shadow cell is used to highlight those values that are close to 95\% confidence.  It should be highlighted that there was not a single question where the proposed tool obtained worse results than the initial team activity. The proposed tool always offered answers that were at least as good as the initial setting, and, in many cases, the responses were statistically more satisfactory. Out of the 15 answers gathered, 9 questions obtain statistically better results than the initial activity by the last iteration of our proposed method (activity 3). One additional question obtained results that are close to being statistically significant. In the first iteration of our proposed tool (activity 2), 8 questions obtained 
statistically better results than the initial activity, and one additional question was close to being statistically significant. Hence, even in intermediate iterations, the proposed tool is capable of obtaining better responses from students. We comment on these results in more detail below.

\begin{table}
\caption{The results of the questionnaire for the three team activities. D: disagree, N: neutral, A: agree}
\begin{center}
\begin{tabular}{ | l l | l | l | l | } 
\hline
& & Act. 1 & Act. 2 & Act. 3 \\
\hline
 \multirow{10}{*}{Team dynamics} & Q1 & D:12, A:36 & \cellcolor{green!25}{\textbf{D:4, A:40 }}& \cellcolor{green!25}{\textbf{D:3, A:37}} \\
 & Q2 & D:15, A:33 & \cellcolor{green!25}{\textbf{D:2: A:42}} & \cellcolor{green!25}{\textbf{D:4, A:36}} \\
 & Q3 & D:11, A:37 & \cellcolor{green!25}{\textbf{D:2, A:42}} & \cellcolor{green!25}{\textbf{D:3, A:37}} \\
 & Q4 & D:16, A:32 & \cellcolor{green!25}{\textbf{D:6, A:38}} & \cellcolor{blue!15}{\textbf{D:8, A:32}} \\
 & Q5 & D:11, A:37 & \cellcolor{blue!15}{\textbf{D:5, A:39}} &  \cellcolor{green!25}{\textbf{D:3, A:37}} \\
 & Q6 & D:16, A:32 & \cellcolor{blue!15}{\textbf{D:8, A:36}} & \cellcolor{green!25}{\textbf{D:4, A:36}} \\
 & Q7 & D:8, A:40 & D:5, A:39 & D:4, A:36 \\
 & Q8 & D:7, A:41 & D:6, A:38 & D:3, A:37 \\
 & Q9 & D:25, A:23 & D:25, A:19 & D:26, A:14 \\
 & Q10 & D:35, A:13 & D:33, A:11 & D:27, A:13 \\
 \hline
\multirow{5}{*}{Student satisfaction} & Q11 & D:7, A:41 & \cellcolor{green!25}{\textbf{D:0, A:44}} & \cellcolor{green!25}{\textbf{D:1, A:39}} \\
 & Q12 & D:18, A:30 & \cellcolor{green!25}{\textbf{D:5, A:39}} & \cellcolor{green!25}{\textbf{D:5, A:35}} \\
 & Q13 & D:7, A:41 & D:2, A:42 & D:3, A:37 \\
 & Q14 & D:10, A:38 & \cellcolor{green!25}{\textbf{D:1, A:43}} & \cellcolor{green!25}{\textbf{D:3, A:37}} \\
 & Q15 & 8.05 & 8.22 & \cellcolor{green!25}{\textbf{8.41}} \\
 \hline
\end{tabular}
\end{center}
\label{tab:result}
\end{table}

\subsubsection{Team dynamics}
The results for the questions related to team dynamics supported the good performance of our proposed tool. Of the 10 questions concerning team dynamics, 5 showed statistically better answers in the third activity (Q1, Q2, Q3, Q5, Q6) than those obtained in the initial activity, and 1 extra question was close to obtaining statistically better answers in the third activity (Q4). Hence, at the end of the experiment, the proposed tool was capable of improving 5 out of the 10 team dynamics  without decreasing the quality of the other 5 team dynamics. In the second activity (which represents an intermediate iteration of the proposed tool), it was also capable of improving 4 out of the 10 team dynamics (Q1, Q2, Q3, Q4) without decreasing the quality of the other 5 team dynamics. Additionally, another question was close to achieving statistical significance (Q5,Q6). 

\begin{figure}[t]
\begin{center}
\includegraphics[width=0.495\linewidth]{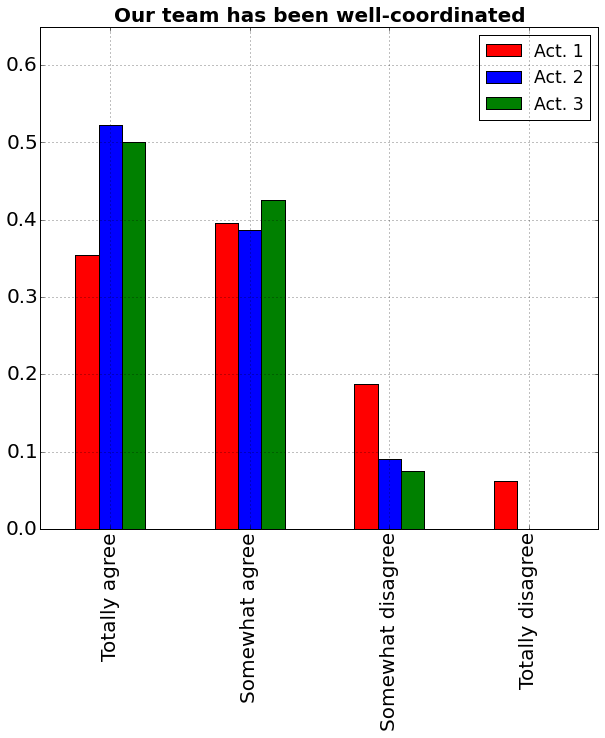}
\includegraphics[width=0.495\linewidth]{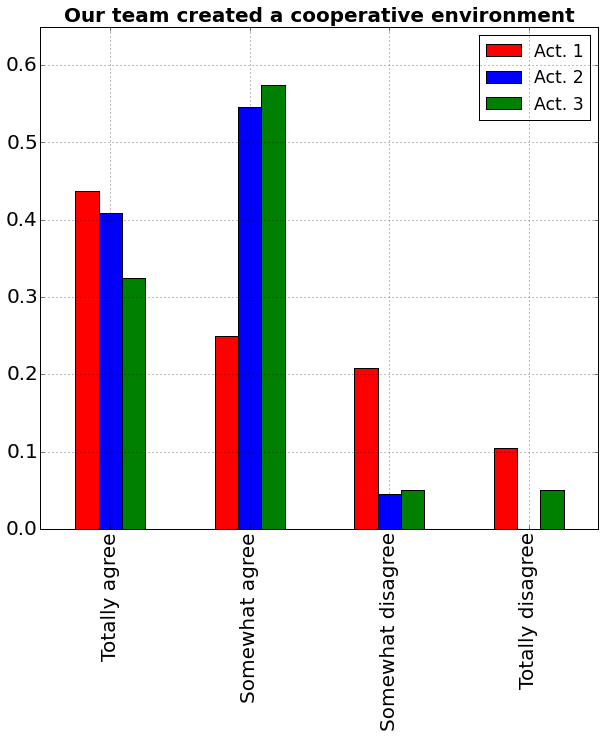}
\caption{Detailed answers for Q1 (left), and Q2 (right)}
\label{img:Q1Q2}
\end{center}
\end{figure}

The first set of questions addresses issues that are related to cooperation and coordination (Q1, Q2, Q3, Q4, Q5), which are critical team performance factors \cite{mathieu00,salas08}. First of all, a higher percentage of students perceived a cooperative environment in those teams proposed by the tool (Q2). On the one hand, an adjusted mid p-value of $0.008$ and the ratio of students that answered this question positively increased by 26 (from 69\% to 95\%)  if we compare the first and the second team activities. On the other hand, an adjusted mid p-value of $0.008$ and an increase of 21 in the ratio of students that answered this question positively (from 69\% to 90\%) was obtained when comparing the first and the third team activities. This shared sense of cooperation was also accompanied by a clearer perception of good coordination among team members (Q1). In this case, if we compare the first and the second team activities, the adjusted mid p-value was $0.02$ with an increase of 15\% in the ratio of 
students that that answered this question positively (from 75\% to 90\%). When compared with the first team activity, the third team activity obtained an adjusted mid p-value of $0.02$ with an increase of 17\% in the ratio of students that answered this question positively (from 75\% to 93\%). A detailed view on how students answered these two questions is shown in Fig. \ref{img:Q1Q2}. As can be observed, the ratio of students that answered these both questions positively for the second and third team activities is larger than the ratio of students that answered these both questions positively on the first team activity.

 It is likely that good coordination and cooperation were possible due to the fact that those teams proposed by the tool were able to more frequently agree on clear team norms and commitments (Q3, Q4), and time was managed in a more efficient way (Q5). In the case of Q3, the adjusted mid p-values obtained when comparing the second and the third team activities with the first team activity were $0.01$ and $0.02$, respectively, whereas there was an increase of 18 and 16 in the ratio of students that answered this question positively (from 77\% to 95\% and 93\%, respectively). The results for question Q4 show an adjusted mid p-value of $0.02$ when comparing the first and the second team activities with an increase of 20 in the ratio of students that answered this question positively (from 66\% to 86\%). When comparing the first and the third activities, the adjusted p-value was found to be $0.08$ and we cannot conclude that students considered that all of the teammates did their share of the work. 
Despite this, the mid p-value is very close to statistical significance. Hence, we consider that the result for this question is partially supported. Nevertheless, we observed an increase of 14 in the ratio of students that answered this question positively (from 66\% to 80\%) when comparing the first and the third team activities. For Q5, we could not find a significant statistical difference between the first and the second team activities, with an adjusted mid p-value of $0.07$. However, we did observe an increase in the ratio of students that answered this question positively (from 77\% to 88\%). We observed statistical differences between the third and the first team activities. In that case, the adjusted p-value was $0.04$ and there was an increase of 16 in the ratio of students that answered this question positively (from 77\% to 93\%). Hence, taking into account all of the previous findings, it can be observed that the proposed tool is able to form teams that attain higher levels of coordination and 
cooperation (i.e., Q1, Q2, Q3, Q5, and partially Q4).

The second set of questions involves team member behaviours within the team (Q6, Q7). As pointed out by the results, a significant and higher percentage of students perceived a positive attitude in team members when the teams were proposed by our tool (Q6) during the third team activity.  The adjusted mid p-value for this comparison was found to be $0.009$ with an increase of 24 in the ratio of students that answered this question positively (from 66\% to 90\%). This positive attitude may be a direct consequence of (or be reinforced by) the general feeling of cooperation and coordination (Q1, Q2). The comparison between the first and the second team activities obtained an adjusted mid p-value of $0.05$, which only partially supports the finding, and there was an increase of 16 in the ratio of students that answered this question positively (from 66\% to 82\%). Even though the tool was not able to generate teams that perceived a more positive attitude in teammates in the second activity, after the third 
iteration, the algorithm was capable of providing a positive environment with the teams that were generated. The reason for this may be explained by the fact that the algorithm has more information with which to provide appropriate team allocations. 

Even though team members had a more positive attitude, we could not find any significant statistical differences regarding how students treated their peers throughout the experiment (Q7). The comparison between the first and the second activities obtained an adjusted p-value of $0.24$, and the comparison between the first and the third activities also obtained an adjusted p-value of $0.24$ too. However, there was an increase in the ratio of students that answered this question positively in the second and third activities with respect to the first team activity (from 80\% to 88\% and 90\%). In all of the cases, the students were generally respectful with others and there was no serious incidents during the experiment.

When it comes to decision-making aspects (Q8, Q9 and Q10), the experiments showed no support for better decision making. Decision-making processes are difficult, especially when they involve groups of individuals. Despite the fact that Belbin's theory has been reported to improve team performance in complex decision-making tasks like management games \cite{prichard99}, we could not find significant improvements over the initial team configuration. In contrast, to other works where Belbin roles have contributed significantly to decision-making tasks \cite{prichard99}, we believe that the main difference between our present scenario and other works resides in the fact that decision making is not the core activity of the projects proposed to students. Naturally, there is always a decision-making component that is inherent to every team activity, but it is perhaps not as strong as in management games. Consequently, it is our belief that differences in decision-making processes are more difficult to observe.

\subsubsection{Student satisfaction}
\begin{figure}[t]
\begin{center}
\includegraphics[width=0.5\linewidth]{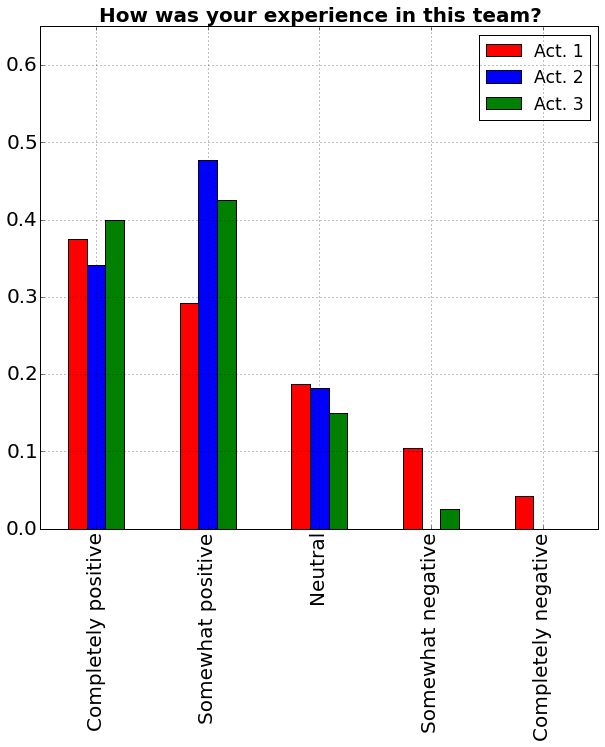}
\caption{Detailed answer for Q11}
\label{img:Q11}
\end{center}
\end{figure}

As Table \ref{tab:result} shows, in the last iteration, the proposed tool is able to obtain statistically better answers to 4 of the 5 questions concerning student satisfaction. More specifically, a statistically higher percentage of students evaluated the team experience as positive and felt more comfortable working on the teams proposed by our team formation tool (Q11, Q12), even in the early stages. In the case of Q11, we observed adjusted mid p-values of $0.008$ and $0.02$, and an increase of 16 and 16 in the ratio of students that perceived positive aspects in the team experience (from 66\% to 82\%) when comparing the first activity with the second and the third team activities, respectively. For Q12, it was observed that adjusted mid p-values strongly supported the use of our tool in the second and third activities with respect to the classic team formation mechanism (adjusted mid p-values of $0.004$ and $0.004$, respectively). We also observed a dramatic increase in the ratio of students that answered 
this question positively (from 62\% to 88\% and 87\%).  A more detailed view of student satisfaction with the team activities can be found in Figure \ref{img:Q11}. As in other figures, the reader can observe how the number of students who answered this question positively is larger for the second and third team activities than for the first team activity.

This is also partially reflected in the evaluations given by students for their respective teams (Q15). Also note that since the tool incorporates more feedback from the students with each activity, the average team evaluation gets higher. Therefore, in the third team activity, the result obtained for Q15 is statistically significant (p-value of $0.04$) and better than that for the first team activity (an increase of 0.4 on a scale of 0 to 10). 

We cannot confirm that a higher percentage of students perceived that they contributed to the team (Q13). Adjusted mid p-values were not close to the $\alpha$ value of significance (0.12 and 0.16 for the second and third activities, respectively) despite the fact that we found increases in the ratios of students that answered this question positively (from 85\% to 95\% and 93\%). In none of the activities did the students feel that they had not contributed to the team. This result should be treated carefully since students may be biased when self-assessing their own contributions \cite{ross06}.

Not only were the students more satisfied with the team, but a higher percentage of students perceived that the work carried out in the group was satisfying for other members on the team (Q14). In this case, the comparison between the first and the second team activities obtained an adjusted mid p-value of $0.006$ with an increase of 19 in the ratio of students that answered this question positively (from 79\% to 98\%). The comparison between the first and the third team activities also offered positive results with an adjusted mid p-value of 0.04 and an increase of 14 in the ratio of students that answered this question positively (from 79\% to 93\%). Therefore, if previous findings are considered in this experiment, the students perceived a higher degree of individual and peer satisfaction.

\subsection{Other findings}
The data that we obtained from the experiments allowed us to explore other issues which are indirectly related to the proposed tool's performance. We analyzed the role distribution in the classroom based on the feedback received from teammates, and considering the most voted role as the most primary role of a student and the second most voted role as the secondary role, . We would like to clarify that this role distribution is by no means generalizable because students from different academic backgrounds (e.g., art, engineering, etc.) may be different and thus play different roles on teams. Nevertheless, we believe that this role distribution may offer insights for future research.

\begin{figure}[t]
\begin{center}
\includegraphics[width=0.495\linewidth]{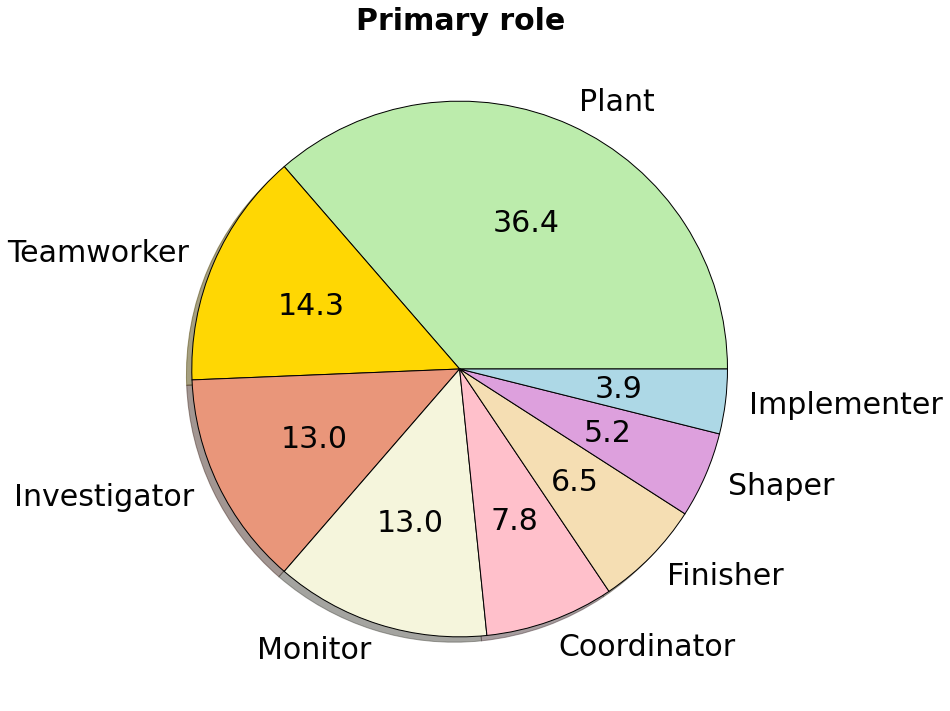}
\includegraphics[width=0.45\linewidth]{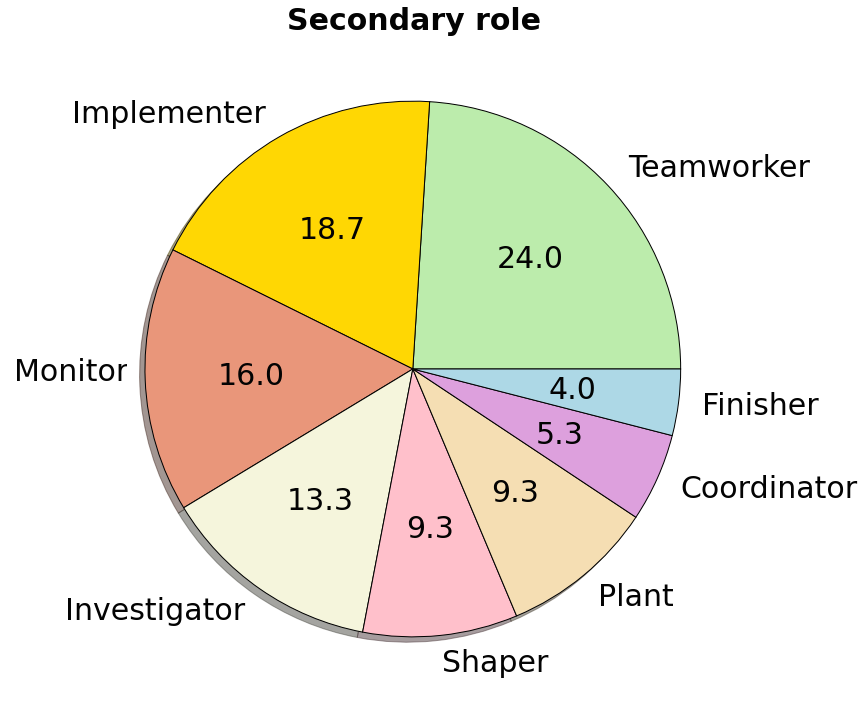}
\caption{Distribution of the primary (left) and secondary (right) roles of the students who took part in the experiment}
\label{img:role}
\end{center}
\end{figure}

Figure \ref{img:role} shows the distribution of roles for the students that participated in the study. The results suggest that students perceived the plant, monitor, team worker, and resource investigator to be the most prominent roles in the classroom, accounting for almost 77\% of the students. This first look at the role distribution clearly indicates that roles are not equally distributed among students. Thinking roles (i.e., monitor and plant) accounted for 49.4\% of the student population in our experiment,  social roles (i.e., coordinator, investigator, and teamworker)  accounted for 35.1\% of the students, and action roles (i.e., shaper, finisher, and implementer) only accounted for 15.6\% of our student population. 

As stated by Belbin's theory \cite{belbin96}, individuals do not play just a single role on a team; they play several roles as needed. A closer look at the second most prominent or secondary role indicates that one of the most frequent secondary roles for students was implementer, which is an action-oriented role. Hence, action-oriented roles may not be as misrepresented as one might initially think. Given the way that our tool computes its optimality function, which takes into account uncertainty and considers as many role allocations as possible (see Equation \ref{expectedteam}), the proposed algorithm is able to cope with situations where individuals are not described by single roles. 

For the vote distribution for individual roles, we observed that the most prominent role for each student on average received 43\% of the votes, while the second most prominente role received 24\% of the votes. These results are certainly far from a random vote, where each role would receive 12.5\% of the votes. This means that there is a certain consensus among students with regard to peer roles: the most prominent role has almost four times more support than a randomly selected role, whereas the secondary role has almost twice the support than random voting would have. 
 
In previous simulations that we carried out to assess the viability of the proposed tool \cite{delval14}, the tool was even robust to human voting with a high level of noise. As stated above, there is a certain level of consensus with regard to the most prominent roles for students. Therefore, in this situation, it is only logical for the proposed tool to be able to detect the most probable roles for students and act accordingly. 
 
We were also able to analyze this behaviour in our data. Of the 34 students that participated in all of the team activities, 11 students had their most prominent role converge after the first team activity's feedback, and it was maintained throughout the experiment. An additional set of 12 students had their most prominent role converge after the feedback from the second team activity. Therefore, almost 68\% of the students had their most prominent role converge in just two team activities. We claim that this is a positive result, especially if we consider that the tool has been designed to share information across different modules. Therefore, the most prominent role of students is available after a few activities.
 
%\label{sec:related}
\section{Related work}\label{sec:related}
The study of the composition and performance of collaborative teams is a topic that has attracted the attention of many researchers for years \cite{Bantel,Jackson,DeDreu,Ensley,Mathieu2008,Andre,Spoelstra201511}. %%2: Collaborative teams have an special interest in learning environments, in which students must work together in small groups in order to solve some task or project. ACÍ buscar algunes referències que gasten per agrupar.
Collaborative learning is widely accepted as a particularly important factor that enhances the learning process in educational contexts \cite{Francescato,Dewiyanti,Tolmie,Schneider,Blasco-Arcas}. However, there are not very many studies that focus on the key issue of team formation, which is a task that requires a considerable amount of time to compose well-balanced teams. %(Moreno et al., 2012)[Yannibelli & Amandi, 2012] BAIS 2013. .
More specifically, when there is a large number of students and different grouping criteria, the task of forming collaborative learning teams to promote successful outputs is considered an exponential problem. In line with this, several computational-based approaches have been proposed in the last few years to provide support to automatically, efficiently, and effectively deal with this goal.

%%%3: Dir que en els últims anys han aparegut ferramentes computacionals per ajudar a la formació de grups, LA MAJORIA D'ELLES TENINT EN COMPTE característiques

The majority of these proposals are oriented to creating heterogeneous teams since there is an extended view in the literature of a direct relationship between the performance of a team and the heterogeneity of its team members. As an example, Graf et al. \cite{Graf2006} present an Ant Colony Optimization approach that provides heterogeneous groups based on the personal traits of students. The groups are made up of by four students with low, average, and high student scores. The algorithm tries to maximize the diversity of the group while keeping a similar degree of heterogeneity in all of the groups. 

Genetic or evolutionary algorithms are also commonly used to solve the problem of forming collaborative learning teams. Wang et al. \cite{Wang2007} present an approach for automatic team formation based on thinking styles \cite{Grigorenko1997} to determine the features of the students. The algorithm translates the features of the students into points in a two-dimensional space which are then classified into categories. The algorithm uses a genetic algorithm to create the optimal group formation based on the categories of the students. %--The experiments consider 66 students and groups of 3 people. At the end of the process, the coalitions are evaluated by the students through a questionary. However, this evaluation is not used as feedback to improve the group formation.--% AÇÒ SERIA ALGO APLICABLE A TOTS, que es podria dir al final

Bergey and King \cite{Bergey14} present a decision-support tool to create heterogeneous student teams based on genetic algorithm with grouping criteria consists of characteristics such as work experience, personality, demographics, undergraduate degree, and academic performance. This tool was applied in university contexts, and the performance of these optimally balanced teams was better than those created manually. %Encara que els grups formats per  la ferramenta van funcionar millor que els manuals, esta ferramenta no gasta el model de Belbin, i a més, només forma grups inicials. TAMBÉ REQUERIX DE LA INTERACCIÓ DEL PROFE??

Lin et al. \cite{Lin2010} present a system that assists instructors in forming collaborative learning groups. The algorithm considers two criteria: information about the understanding levels, and the interests of the students. Particle swarm optimization is used in the group composition algorithm to deal with the complexity of the problem. Even though this proposal presents an algorithm that finds solutions within a reasonable time, the validation of this proposal only focuses on the computational side. Thus, the grouping criteria has not been tested in a real environment in order to measure whether or not the performance of the teams improved. %no ho aplica en entorns reals

Cavanaugh et al. \cite{Cavanaugh04} present a web-based system to create teams that takes into account characteristics such as gender, skills, and student schedules. Moreno et al. \cite{Moreno} provide an approach with underlying genetic algorithms for inter-homogeneous and intra-heterogeneous team formation. In contrast with the above approaches, this proposal does not limit team formation to a specific number of student characteristics.

Another type of algorithm is used by Christodoulopoulos et al. \cite{Christodoulopoulos2007}, who presents a web-based group formation tool that facilitates the creation of homogeneous and heterogeneous groups based on the learning styles of the students. This tool allows the instructor to manually modify the groups and allows the students to negotiate the grouping. The creation of homogeneous groups is based on a Fuzzy C-Means algorithm, and the creation of heterogeneous groups is based on a random selection algorithm. The tool also provides an option to negotiate the proposed teams with the students. However, this negotiation consists of direct interaction with the teacher. %Other approaches use bio-inspired algorithms. 

The problem of team formation is also present in the context of human resource management. As an example, Wi et al. \cite{Wi2009} present a framework to deal with team formation in R\&D-oriented institutes. The authors propose a genetic algorithm that uses a fuzzy model to take into account information about the candidates' knowledge and expertise about certain topics related to a project. The algorithm also takes into account information about the position of the candidates in a social network in order to determine their suitability for project management positions.

As can be observed, there is a large number of computational-based approaches for automatically forming collaborative learning teams that are focused on the individual characteristics of the students (e.g. personality, learning style, learning achievement, or thinking style). These characteristics motivate and enhance the capability of each individual to occupy different team roles, considering a role as the set of goal-directed behaviours taken on by a person within a specific situation \cite{Stewart,aritzeta07,Mathieu2015}.

Even though team roles are considered to be a critical part of making effective teams \cite{Ancona,Belbin1993,Sundstrom,Fransen,Wang2007} and there are several taxonomies of team roles in the literature \cite{Bales,Belbin1993,Belbin2011,Margerison,Parker,Humphrey}, there are very few computational-based approaches that focus on this topic.

%4: algun model de rols. Dir que la majoria de solucions automàtiques tracten característiques dels estudiants, però que en la literatura de psicologia se li dóna molta importància també a models basats en rols, com tal o qual.

In \cite{Bais13} and \cite{Balmaceda}, the authors propose an approach that uses an underlying model proposed by Mumma \cite{Mumma}, which is based on team roles. This approach considers team formation to be a constraint satisfaction problem. Even though the formation tool is applied in a real context, the experiments focus only on the performance of the tool for finding the groups. Thus, the real impact of team performance is not analyzed. %(es decir, si los modelos teóricos realmente se cumplen en la realidad)

%5: Belbin
In the literature, one of the most widely used frameworks for team formation is the Belbin model \cite{Belbin1993,Belbin2011}. This model has been widely applied in different contexts such as organizations, firms, and executive education programs \cite{Sommerville,prichard99,Park,Jefferies,aritzeta07,McDonald}. The Team Role Self-Perception Inventory, which is filled out by each student, is the most frequently used tool to collect the information of each individual \cite{Sommerville,fisher96,rajendran05,senior98,furnham93}. However, there are some studies that also consider the Observer Assessment Sheet \cite{senior98,Broucek}. 

Despite the popularity of the Belbin model, there are only a few computational-based approaches that focus on team formation based on this model. One of these approaches was proposed by Yannibelli and Amandi \cite{yannibelli2012a,yannibelli2012b}. It is based on hybrid evolutionary algorithms to find the most suitable collaborative learning teams in the context of software engineering modules. The grouping criterion is defined on the basis of the Belbin model and uses the Self-Perception Inventory. Thus, the algorithm balances the roles and the number of members. The validation of this proposal is based on computational experiments. Therefore, although the performance of the algorithms is promising, the effectiveness of the approach in a real context was not tested.

Fares and Costaguta \cite{Fares} present a multi-agent system that promotes team roles in groups in order to achieve successful learning. This proposal uses interface agents to diagnose the state of the collaboration of each individual within the group, in order to determine each of the roles proposed by Belbin. Similarly to the previous approach, this proposal was not validated in a real context.

%Batenburg et al. (2013) també parla de resultats ambiguos...

In comparison with the above approaches, our proposal presents three different contributions that go a step further into the state of the art. First, our proposal contributes to the development of computational solutions for the team formation  problem, which is based Belbin's model. In \cite{alberola}, we evaluated our proposal with synthetic data in order to test the feasibility of the tool. However, to validate our proposal, we decided to test the performance of the tool in a real experiment involving students, which corresponds to the results presented in this article. Even though the time performance of the algorithm is important, we do not consider the computational cost in this domain to be critical because the task of team formation is only carried out a few times each academic year. Thus, our validation focuses on showing the quality of the team formation model in a real scenario and testing the impact of applying the Belbin model for team formation. Actually, as observed in the real experiments, 
the majority of the students have a prominent role that converges in two team activities. This conclusion is very similar to the convergence rate obtained with synthetic data.

The second key contribution of our proposal is related to the information used as the input for the team formation process. In all of the previous approaches, the information of each student is collected prior to working on a team. More specifically, the Self-Perception Inventory (or other psychological tests used in different models) are used by other approaches to establish the role of each individual. This role assignment is not expected to change and, therefore, the composition of the team would be similar if several iterations  were carried out with the same population of students. However, the behaviour of an individual in a real team environment may be different than one's own perception since moods, emotions, self-esteem and trust can change during the teamwork experience \cite{senior98, Ku, Pfaff, Jones}. In our proposal, each student is evaluated by his/her team partners after the development of some joint project or activity. Thus, the role profile considers the aggregated opinion of the rest of 
the students who worked  with him/her. Since the evaluation is carried out after exhibiting the behaviour in a real situation, the role assignment may be more reliable and precise. What is more, the collected information increases as the number of teamwork projects increases.

Although some of the above approaches include student feedback as a measure of satisfaction, student feedback is not used to improve team formation. Similarly, non-computational studies that apply the Observer Assessment Sheet do not consider that the information retrieved may be different depending on the team activity and that individuals are required to know each other well (which is not always feasible in higher education contexts). This led us to the third key contribution of our proposal, which is related to how roles are assigned to individuals. Since roles are established by more than just an initial questionnaire, the role assigned to each individual is not necessarily limited to a single strict role. In contrast to other approaches, we use probabilities to establish how roles are played by each student. Therefore, we do not establish that an individual strictly plays a predominant role, but we establish that an individual plays each of the eight roles with a specific associated probability. This 
allows us to account for the fact that humans display a variety of behaviors in real situations.

It must be noted that our proposal does not need previous information about the abilities, attributes, or roles played by the students. The information used to characterize the role of each individual is gathered and grows each time a different project or activity is carried out. To our knowledge, our work is the first one that provides this aspect; this improves the information that is collected from the students and establishes a more reliable role assignment since it considers the opinion of other members instead of self-perception. 

%Therefore, the feedback provided by the students is not only used as a satisfaction measurement but it is also used to improve the performance of the team compositions in the following projects or activities\footnote{llevaria esta frase}.

%\label{sec:con}
\section{Conclusions}\label{sec:con}
In this article, we have presented an artificial intelligence tool for team formation in the classroom. The tool forms teams based on Belbin's role theory, which identifies eight different behavioural patterns (i.e., roles) that are found in successful teams. As proposed by the theory, teams are formed with a heterogeneous configuration of roles. These roles are inferred from feedback that is received from students with respect to their teammates after each team activity. Artificial intelligence allows us to handle two important issues for the tool: (i) it can handle uncertainty with respect to individuals' roles and students' feedback by means of Bayesian techniques; (ii) it allows us to calculate the optimal team configuration for the classroom in an enormous search space by using coalitional structure generation approaches that are supported by linear programming.

We tested the performance of the proposed tool in an experiment involving students that took part in three different team activities. The experiments suggest that the proposed tool is able to improve different teamwork aspects such as team dynamics and student satisfaction. 

On the one hand, the results show that students perceived a higher degree of cooperation and coordination in teams proposed by the tool than those teams formed by classical methods. On the other hand, the experiment also provides greater support for students to perceive their teammates' attitudes as being positive. Despite these positive findings and the studies found in the research literature \cite{prichard99}, we could not identify statistical differences with respect to decision-making tasks. We argue that this result may be influenced by the fact that the team activities were not oriented towards decision-making tasks and, therefore, differences in decision making may be more difficult to identify.

Finally, the experiments also suggest that the teams formed by our tool received more positive answers regarding student satisfaction aspects. In general, a higher percentage of students evaluated the team experience as positive, felt more comfortable working with teammates, perceived the work carried out in the team as satisfactory for all teammates, and evaluated their teammates higher.

We must note that although we used a random configuration for the initial team formation, we could include any questionnaire in order to have previous information about each student. In addition, the modularity of our proposal would allow us to use any other underlying grouping model such as another role-based model in order to compare its effectiveness in educational contexts. We plan to deal with this issue in future works.

As for future lines of research, our current tool takes into account the behaviours of individuals on a team. However, there are other factors that we believe will have an impact on team performance. We consider interpersonal relationships to be a factor to be considered in future lines of work since they may affect team dynamics \cite{myers85}. We also plan to include other types of factors such as student performance, level of knowledge, as well as constraints imposed by teachers. 

\section*{Acknowledgements}
This work is supported by the following projects: TIN2014-55206-R, TIN2012-36586-C03-01, PROMETEOII/2013/019, TIN2015-65515-C4-1-R.

\section*{References}
\bibliographystyle{elsarticle-num} 
\bibliography{article}

\end{document}